\documentclass{article}
\usepackage{spconf,amsmath,graphicx}
\usepackage{amsmath}
\usepackage{amsfonts}
\usepackage{graphicx}
\usepackage[table,xcdraw]{xcolor}

\usepackage{booktabs}
\usepackage{enumitem}
\usepackage{multirow}
\usepackage{hyperref}
\usepackage{tikz}
\usetikzlibrary{calc}
\usetikzlibrary{shapes.geometric}

\DeclareMathOperator*{\argmin}{\arg\mkern-2mu\min}

\newcommand{\norm}[1]{\lVert #1\rVert}

\title{MelHuBERT: A Simplified HuBERT on Mel Spectrograms}

\name{Tzu-Quan Lin$^1$, Hung-yi Lee$^1$, Hao Tang$^2$}
\address{
  $^1$Graduate Institute of Communication Engineering, National Taiwan University, Taiwan\\
  $^2$University of Edinburgh, United Kingdom
}

\copyrightnotice{979-8-3503-0689-7/23/\$31.00~\copyright2023 IEEE}
\begin{document}

\maketitle
 
\begin{abstract}
Self-supervised models have had great success in learning speech representations
that can generalize to various downstream tasks.
However, most self-supervised models require a large
amount of compute and multiple GPUs to train, significantly hampering
the development of self-supervised learning.
In an attempt to reduce the computation of training,
we revisit the training of HuBERT, a highly successful self-supervised model.
We improve and simplify several key components, including the loss function, input representation,
and training in multiple stages.
Our model, MelHuBERT,
is able to achieve favorable performance
on phone recognition, speaker identification, and automatic speech recognition against HuBERT,
while saving 31.2\% of the pre-training time, or equivalently 33.5\% MACs per one second speech.
The code and pre-trained models are available in \url{https://github.com/nervjack2/MelHuBERT}.

\end{abstract}

\section{Introduction}

Self-supervised models have shown to be able to generalize across various downstream tasks,
either as feature extractors or through fine-tuning
\cite{chung2019unsupervised, schneider2019wav2vec, baevski2020wav2vec, chung2020vector, hsu2021hubert, chen2022wavlm, huang2022masked}.
To make use of large quantities of unlabeled data, self-supervised models
are often large in terms of both the number of parameters
and the amount of compute needed during inference \cite{zhang2022bigssl}.
One could argue that the amortized cost of training
large models on large data sets is low, because
the resulting models can be reused many times and across
many downstream tasks.
However, the ever increasing cost to train large self-supervised models
on large data sets hinders the development of better training recipes.
In fact, few can afford the cost to train
their own self-supervised models from random initialization.
In this work, we tackle this problem by studying and simplifying
a training recipe of self-supervised models.

Our model of choice is HuBERT \cite{hsu2021hubert}, a self-supervised model
that performs well but also has a simpler training recipe than others.
The training of HuBERT consists of stages.
Each stage performs so-called masked prediction,
a frame-wise prediction task with the input partially masked
\cite{baevski2020wav2vec, hsu2021hubert, chen2022wavlm}.
The stages differ in the targets used for pre-training.
The targets of the first stage are cluster assignments of MFCCs frames (with $k$-means),
while the targets of subsequent stages are cluster assignments of HuBERT hidden layers.
Despite having a simple training recipe, some design decisions
in HuBERT are derived from wav2vec 2.0 without much justification.
It is also generally unclear how the design decisions impact the performance on downstream tasks.
In this work, we identify and study several key components, such as the loss function,
the number of target clusters, the impact of multiple states, the input frame rate,
and the architectural choices.

A simple analysis shows that the first seven convolution layers in HuBERT
constitute 33\% of multiply-accumulation operations (MACs) of the entire
forward computation (including the 12-layer Transformer).
This result motivates us to replace convolution layers with commonly used
speech features, such as Mel spectrograms.
In fact, much prior work has been based on Mel spectrograms \cite{chung2019unsupervised, jiang2019improving, ling2020decoar, huang2022masked, baade2022mae, chong2023masked}.
Since replacing convolution layers with Mel spectrograms
is a significant change to HuBERT,
we term our model MelHuBERT.

A fundamental question is how the choice of input representation
impacts the learned representations.
For many linguistic and paralinguistic properties, e.g., for the use
of automatic speech recognition (ASR) and speaker identification (SID),
Mel spectrograms should be sufficient \cite{sainath2015learning, ravanelli2018speaker}.
In this work, we thoroughly compare speech representations learned by HuBERT and MelHuBERT
on phonetic recognition, speaker identification, and automatic speech recognition.
We will show that MelHuBERT performs favorably over HuBERT when pre-trained
on the 360-hour subset of LibriSpeech, while saving 31.2\% of the training time.

The fact that convolution layers can be replaced by Mel spectrograms
does not imply that the two are equivalent in representation.
We further conduct analyses with canonical correlation analysis (CCA)
to show the degree of similarities between the convolution layers in HuBERT and Mel spectrograms.
We are also able to identify the differences and strengths of the two.

We will begin with a review of HuBERT,
discuss our changes to the original design,
and present the experiments and analyses.

\section{Overview of HuBERT}

HuBERT takes waveform as input and has 7 convolution layers 
to process waveform into frames of 25 ms context and 20 ms shift.
After the convolution layers, there is a 12-layer Transformer,
predicting quantized MFCC frames.
More formally, suppose we have an input utterance $u$ represented
as a sequence of wave samples,
and $y_1, \dots, y_T$ is a sequence of the corresponding 10-ms MFCC frames.
We first quantize $y_1, \dots, y_T$ to a sequence of indices $c_1, \dots, c_T$.
Specifically, $c_t = \argmin_{c = 1, \dots, k} \norm{y_t - v_{c}}_2$,
where $v_c$ is the $c$-th centroid.
We denote $g$ and $f$ as the convolution network and the Transformer,
and denote the outputs of both as
\begin{align}
x_1, \dots, x_{T/2} &= g(u) \\
o_1, \dots, o_{T/2} &= f(x_1, \dots, x_{T/2})
\end{align}
The goal of HuBERT is to minimize
\begin{align}
-\sum_{t=1}^{T/2} \log \frac{\exp(\cos(W o_t, m_{c_{2t}})/\tau)}{
    \sum_{c=1}^k \exp(\cos(W o_t, m_c)/\tau)},
\label{eq:hubert_loss}
\end{align}
where $m_c$ is the embedding of the $c$-th codeword in a learnable codebook $m$,
$W$ is a trainable projection matrix, and $\tau$ is a temperature hyperparameter.

\section{Proposed Measurements}

\begin{figure}[t]
  \centering
  \includegraphics[width=1.0\linewidth]{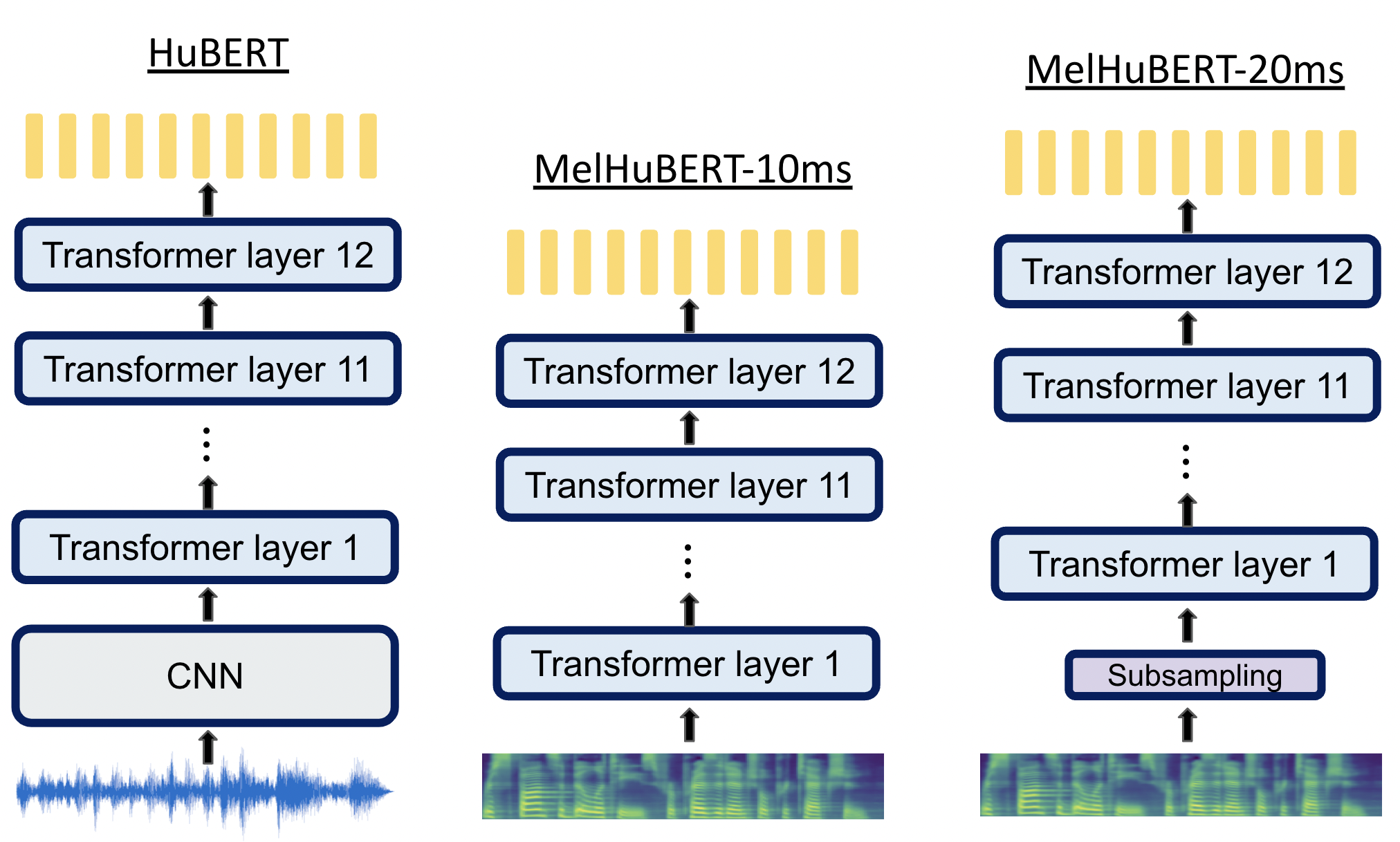}
  \caption{ An overview of HuBERT and MelHuBERT on model architecture.
    \label{fig:melhubert_architecture}}
\end{figure}

Based on the overview of HuBERT, we can already see a few oddities,
for example, the mismatch of frame rates between 10-ms MFCC frames
and the output of convolutional network, and the ad-hoc loss function.
Below we propose several modifications to simplify the training pipeline.

\subsection{Mel spectrograms as input}

Inherited from the original CPC \cite{oord2018representation},
wav2vec 1.0 \cite{schneider2019wav2vec}, wav2vec 2.0 \cite{baevski2020wav2vec},
and HuBERT \cite{hsu2021hubert} all take wave samples as input.
In this work, however, we decide to use Mel spectrograms as input.
Formally, we overload the sequence $x_1, \dots, x_T$ to
denote log Mel features at a 10-ms frame rate,
and pass it directly to the Transformer.
The targets are now quantized log Mel frames,
or equivalently $c_t = \argmin_{c=1, \dots, k} \norm{x_t - v_c}_2$,
without depending on MFCC frames.

There are several reasons why Mel spectrograms are preferred.
For speech tasks, such as automatic speech recognition
and speaker recognition, the benefit of learning
to process wave samples is marginal \cite{sainath2015learning, ravanelli2018speaker}.
In addition, training with Mel spectrograms can achieve
state-of-the-art performance \cite{zhang2022bigssl}.
Finally, despite that wave samples are taken as input in HuBERT,
MFCC frames are still used during training.
More importantly, the convolution network for extracting
acoustic frames actually consumes 33\% of the total MACs
of HuBERT.
We expect that removing the convolution layers
not only saves a significant amount compute, but also
has marginal impact on the learned representation.

\subsection{Cross entropy as the loss function}

The loss in \eqref{eq:hubert_loss} for training HuBERT
is also inherited from CPC \cite{oord2018representation},
wav2vec 1.0 \cite{schneider2019wav2vec}, and wav2vec 2.0 \cite{baevski2020wav2vec}.
The difference, however, is that
the loss in wav2vec 2.0 and the like
is based on noise contrastive estimation \cite{gutmann2010noise}, where
negative samples are required to compute the denominator in the loss.
HuBERT, on the other hand, does not rely on negative samples,
as can be seen in \eqref{eq:hubert_loss}.
This motivates us to simplify the loss function into
\begin{align}
-\sum_{t=1}^T \log \frac{\exp(w_{c_{t}}^\top o_t)}{\sum_{c=1}^k \exp(w_{c}^\top o_t)},
\label{eq:melhubert_loss}
\end{align}
where $c_t = \argmin_{c=1, \dots, k} \norm{x_t - v_c}_2$, i.e., the quantization of 
log Mel features.

The simplified loss function is the familiar cross entropy for predicting
the $k$-means centroids.
In contrast to other generative losses (such as $\ell_2$ on the log Mel features)
used in APC \cite{yang2022autoregressive}, DeCoAR 2.0 \cite{ling2020decoar}, and MPC \cite{jiang2019improving},
this loss function is contrastive.
However, this loss function of predicting the $k$-means of the input also has
a variational interpretation of maximizing the likelihood \cite{yeh2022autoregressive},
making the distinction between generative and contrastive less clear-cut.

\subsection{Subsampling}

If our model takes regular Mel spectrograms as input, the input and output frame rate
would be 10 ms, as opposed to HuBERT's 20 ms.
To ensure that they have the same frame rate, 
we choose to subsample Mel spectrograms,
concatenating every two contiguous frame to double the frame rate.
As shown in Figure \ref{fig:melhubert_architecture},
we use the term MelHuBERT-10ms to refer to the model with a frame rate of 10, 
and MelHuBERT-20ms to refer to the variant with a frame rate of 20.
Due to the difference in input sequence length,
MelHuBERT-10ms runs much slower than its 20ms variant. 
As we will show in Section 4.1, 
MelHuBERT-20ms has smaller MACs than MelHuBERT-10ms.
In addition, MelHuBERT-20ms performs competitively or even better than MelHuBERT-10ms on downstream tasks.

\section{Experiments}

Following prior work \cite{chung2019unsupervised, baevski2019vq}, we pre-train our models on subsets of LibriSpeech and evaluate downstream tasks on respective data sets.

For training MelHuBERT, we use a learning rate of $10^{-4}$ and do not apply any learning rate scheduling during pre-training.
Similar to HuBERT, dropout with the probability of 10\% is applied
after most matrix multiplications, such as after query, key, and value multiplication,
and after FC1 and FC2.
We do not use LayerDrop \cite{fan2019reducing} and mixed precision training \cite{micikevicius2017mixed}.
It takes around 150 hours to train MelHuBERT-20ms for 200 epochs
on a single 24 GB 3090 GPU with an effective batch size of 32
(accumulated 4 times with a batch size of 8).

We train HuBERT using the training script in Fairseq \cite{ott2019fairseq}.
We use an effective batch size to 32, train for 200 epochs, and disable mixed-precision
training, to match the setting of MelHuBERT.
Training is otherwise exactly the same as the official release.  

We follow the protocol of SUPERB \cite{shu2021superb} and S3PRL\footnote{https://github.com/s3prl/s3prl},
computing a weighted sum of all hidden layers as the representation
for downstream tasks.
All self-supervised models are frozen, and the weights for each layer are learned for each task.
We focus on phone recognition (PR), speaker identification (SID), and automatic speech recognition (ASR).
Phone recognition and automatic speech recognition are conducted on the 100-hour subset of LibriSpeech, 
and phone error rate and word error rate are reported respectively.
Speaker identification is conducted on Voxceleb1,
and classification error rate is reported. 
We find that the default learning rates in SUPERB require tuning.
For MelHuBERT, the learning rates for PR, SID and ASR are $10^{-4}$, $10^{-2}$, and $10^{-4}$,
respectively, and for HuBERT, $10^{-3}$, $10^{-1}$, and $10^{-4}$.

\subsection{Stage-1 initial results}

We first present our initial experiments of training
MelHuBERT by simply replacing the
convolution layers with log Mel features and using cross entropy
as the loss function.
Detailed exploration will be discussed in the next section.

The input to MelHuBERT is 40-dimensional Mel spectrograms,
normalized with global mean and variance.
We quantize the same input Mel spectrograms as our targets,
using $k$-means with 512 clusters.
We concatenate every two contiguous frames to produce
the input for MelHuBERT-20ms.
The targets for training MelHuBERT-20ms are the cluster labels of the odd frames.
The downstream performance and MACs
are shown in Table \ref{tbl:initial}.
Similar to the findings in prior work \cite{meng2022compressing},
downsampling the input (by concatenation) does not suffer degradation on downstream tasks.
Overall, MelHuBERT behaves similarly to HuBERT while saving 33.5\% MACs per one second speed for MelHuBERT-20ms.
The results aligns well with prior studies \cite{wu2022performance, peng2023structured, peng2023dphubert, parcollet2023efficiency, chen2023front}, showing that it is possible to prune convolution layers without much performance degradation.
In favor of the faster runtime (in both training and inference), we will focus on MelHuBERT-20ms for the rest of the paper. 

\begin{table}[t]
  \caption{A comparison of MelHuBERT and HuBERT
  (pre-trained on the 360-hour subset of LibriSpeech)
  based on frame period (FP), MACs per one second speech, phone recognition (PR),
  and speaker identification (SID). 
  \label{tbl:initial}}
  \begin{center}
  \scalebox{0.88}{
  \begin{tabular}{ccccccc}
  \toprule
  \multirow{2}{*}{Model} & FP   & MACs  & PR    & SID  & ASR \\
  & (ms) & (G/sec) & (PER) & (ER) & (WER) \\
  \midrule  
  MelHuBERT-10ms & 10 & 10.76 & 15.1 & 35.2 & \textbf{11.6} \\ 
  MelHuBERT-20ms & 20 & 4.93 & \textbf{13.0} & 33.7 & 11.9 \\
  HuBERT & 20  & 7.42 & 13.8 & \textbf{29.7}  & 12.6 \\
  \bottomrule
  \end{tabular}
  }
  \end{center}
\end{table}

\subsection{Stage-1 pre-training}

\begin{table}[t]
  \caption{Results of various design choices, including the number of $k$-means    clusters, the number of Mel filters, and the number of targets
    during pre-training, and the loss functions, for training MelHuBERT-20ms (on 360-hour subset of LibriSpeech).
    \label{tbl:ablation}}
  \begin{center}
  \scalebox{0.85}{
  \begin{tabular}{cccccccc}
  \toprule
  & \multirow{2}{*}{Loss} & \multirow{2}{*}{cluster} & Mel & \multirow{2}{*}{targets} & PR  & SID & ASR \\
  &  &  & filters  &  & (PER) & (ER) & (WER) \\
  \midrule
  1. & Eq \eqref{eq:hubert_loss}     & 512 & 40 & 1  & 13.3  & 34.6  & 12.3   \\
  2. & CE         & 512 & 40 & 1  & 13.0  & 33.7 &  11.9   \\
  3. & CE         & 100 & 40 & 1  & 12.4  & 31.7  & 11.7  \\ 
  4. & CE         & 100 & 80 & 1  & 12.9  & 32.7  & 11.8 \\
  5. & CE         & 512 & 40 & 2  & 12.4  & 32.7  & 11.5 \\
  6. & CE         & 100 & 40 & 2  & \textbf{12.3}  & \textbf{30.5} & \textbf{11.3} \\
  \bottomrule
  \end{tabular}
  }
  \end{center}
\end{table}

The initial experiments MelHuBERT show promising results.
In this section, we further study the individual choices
of $k$-means clusters, the number of Mel bins, and the loss functions.
Results are shown in the Table \ref{tbl:ablation}.

HuBERT's loss function \eqref{eq:hubert_loss} is inherited
from wav2vec \cite{schneider2019wav2vec}
and wav2vec 2.0 \cite{baevski2020wav2vec}.
We choose cross entropy for its simplicity,
and compare the two in this section.
Results are shown in the first two rows of Table \ref{tbl:ablation}.
We see a slight improvement in PR, SID, and ASR when using the cross entropy.

The original study of HuBERT \cite{hsu2021hubert} experiments with 50, 100, and 500 clusters when quantizing MFCCs, and finds that using 100 clusters performs better than 500 clusters on ASR.
The number of clusters is also studied in \cite{yeh2022autoregressive}
in the context of autoregressive predictive coding (APC) with LSTMs,
and similar to the finding of \cite{hsu2021hubert}, 
using 100 clusters performs better than 512 clusters on phone classification.
On the contrary, 512 clusters works best in vector-quantized APC for phone classification \cite{chung2020vector}.
The number of clusters has a different impact on speaker tasks, and is in general less explored \cite{chung2020vector}.
We compare different numbers of clusters in training MelHuBERT.
Results are shown in row 2 and 3 of Table \ref{tbl:ablation}.
We find that using 100 clusters leads to better overall results, especially on speaker identification.

Next, we study the number of Mel filters used to extract Mel spectrograms.
Various numbers of Mel bins have been used in prior work.
For example, APC \cite{chung2019unsupervised, yeh2022autoregressive} uses 40 filters;
Mockingjay \cite{liu2020mockingjay}, DeCoAR 2.0 \cite{ling2020decoar},
and TERA \cite{liu2021tera} use 80 filters; 
while SSAST \cite{gong2022ssast} uses 128 filters.
We explore this option for MelHuBERT, and the results are shown in row 3 and 4 of Table \ref{tbl:ablation}.
We find that using using more Mel filters does not lead to
any improvement.

Instead of only using the cluster labels of the odd frames,
we also explore predicting the cluster labels of both the odd and the even frames.
Specifically, when the two frames at $t$ and $t+1$ are
concatenated, we compare training MelHuBERT-20ms against
only the cluster labels at $t$ to
the one against both $t$ and $t+1$.
We simply use two projection matrices when predicting two targets,
a multitask approach.
Results are shown in row 2 and 5 of Table \ref{tbl:ablation}, and we do find that
using two targets is better than one.

Finally, combining all the previous findings, 
we use cross entropy as the loss function, 
100 clusters as targets, 40 Mel filters, 
and predicting multiple both the odd and the even targets. 
The result of our final stage-1 model is presented in the last row of Table \ref{tbl:ablation}. 
The small adjustments lead to a sizeable improvement over
our initial MelHuBERT.
We will proceed with the best setting
for the rest of the paper.

\subsection{Stage-2 pre-training}

After the first stage\footnote{In \cite{hsu2021hubert}, HuBERT is trained in multiple \emph{iterations}, where the current iteration is trained on targets produced by the model from the previous iteration. We decide to use the term \emph{stages} instead of iterations, because the differences among multiple stages are quite significant.}, HuBERT includes multiple subsequent
stages to train their models \cite{hsu2021hubert}.
We limit ourselves to two stages.
The second stage uses quantized hidden vectors as
targets for training, and we need to decide which
layer to quantize.
In the original study, hidden vectors of each layer
are clustered, and the phone purity and cluster purity
are measured based on forced alignments.
The layer with the highest purity is chosen for stage two,
since it aligns most closely with the phone units.
The effect of multi-stage training is reported in \cite{hsu2021hubert}
with a focus on ASR.
It is unclear how this decision might impact phone recognition and speaker identification.
Following HuBERT, we first calculate the phone purity and cluster purity of hidden layers,
based on force alignments obtained
from the Montreal Forced Aligner \cite{mcauliffe2017montreal}.\footnote{In our opinion,
the pipeline seizes to be unsupervised if it relies on forced alignments.
We include the results and analyses for completeness.}

Similar to \cite{hsu2021hubert,pasad2021layer},
our phone purity and cluster purity result show that the phone information is most accessible in
the middle layers, with the sixth layer being the best, consistent
with \cite{hsu2021hubert}. 
For comparison, 
we have also conducted stage-2 pre-training for HuBERT.
We train MelHuBERT-20ms and HuBERT on the quantized hidden vectors
produced by their respective stage-1 models.
For MelHuBERT, we also explore training the second stage from random initialization and continued pre-training. 
For HuBERT, we train the second stage from random initialization (based on the communication with the authors in \cite{hsu2021hubert}).
Same as the first stage, we train MelHuBERT and HuBERT
with a learning rate of $10^{-4}$ and
an effective batch size of 32 for 200 epochs.

Results comparing stage one and two are shown in Table \ref{tbl:stage-2}.
Both continued pre-training and training from random initialization in stage two yield a significant improvement in PR and ASR.
However, training from scratch performs much better at speaker identification.
Overall, MelHuBERT and HuBERT behave similarly, with MelHuBERT having a slight edge over HuBERT on ASR and SID.
We will provide more in-depth analyses on the differences between stage one and two in the next section.

\begin{table}
  \caption{Results of stage-2 pre-training on PR, SID, and ASR. MelHuBERT-20ms (scratch) is trained
    from random initialization, while MelHuBERT-20ms (cont) continues pre-training from stage 1.
    \label{tbl:stage-2}}
  \vspace{1em}
  \centering
  \scalebox{0.95}{
  \begin{tabular}{ccccc}
  \toprule
                    & stage & PR  & SID & ASR \\
  \midrule
  MelHuBERT-20ms    &  1    &  \textbf{12.3}  & 30.5 & \textbf{11.9} \\
  HuBERT            &  1    &  13.8  & \textbf{29.7} & 12.6 \\
  \midrule
  MelHuBERT-20ms (scratch)    &  2    &   8.6   &  \textbf{25.3}  & \textbf{9.7} \\
  MelHuBERT-20ms (cont)    &  2    &   8.7   &  31.7  & 9.8 \\
  HuBERT            &  2    &   \textbf{8.1}    &  26.6  & 10.2\\
  \bottomrule
  \end{tabular}
  }
\end{table}

\section{Analysis}

Given that MelHuBERT and HuBERT perform similarly on downstream tasks,
there are several questions remained open.
Since MelHuBERT and HuBERT are self-supervised models for learning
speech representations,
the first and the most important question is whether the two
learn different representations.
If the two do learn different representations, the question
becomes whether we can characterize the differences.
Finally, if we can characterize the differences,
what their strengths and weaknesses are is also worth studying.
In this section, we conduct a set of experiments to
answer these questions.

\subsection{Layer-wise similarity to phones}

Since we replace seven convolution layers with Mel spectrograms,
the immediate question is how it impacts the representation learned.
We will focus on the phonetic information, as it is
important for phone recognition and ASR.
The analysis comparing the first and the second stage of pre-training
is also generally lacking in prior work.
To answer both questions,
we study how HuBERT differs from MelHuBERT with layer-wise analysis
proposed by \cite{pasad2021layer, pasad2023comparative}
and compute the CCA similarity between one-hot vectors of phones
and mean-pooled phone-level representations.

The analysis of phonetic information across layers is shown in the top plot of Figure \ref{fig:cca_layerwise}. 
HuBERT and MelHuBERT behave similarly, and there is no significant difference from the first Transformer layer to the ninth.
Compared to HuBERT, MelHuBERT has a more significant drop in phone similarity before the first
and after the tenth Transformer layer.
This can be attributed to the autoencoding behavior, and is consistent with
the observation in \cite{pasad2021layer}.

Compared to stage 1, the phone similarity of stage 2 (for both HuBERT and MelHuBERT)
is higher after the eighth Transformer layer and stays high.
This can be attributed to the targets for stage 2 training,
because the targets are from the sixth layer of stage 1 and
are phonetically more prominent.

\subsection{Layer-wise similarity to Mel spectrograms}

The fact that HuBERT and MelHuBERT has similar downstream performance
does not necessarily imply that the convolution layers are computing features similar to
Mel spectrograms.
To study whether the two share anything in common, we compare similarity
to Mel spectrograms for all layers in HuBERT and MelHuBERT.

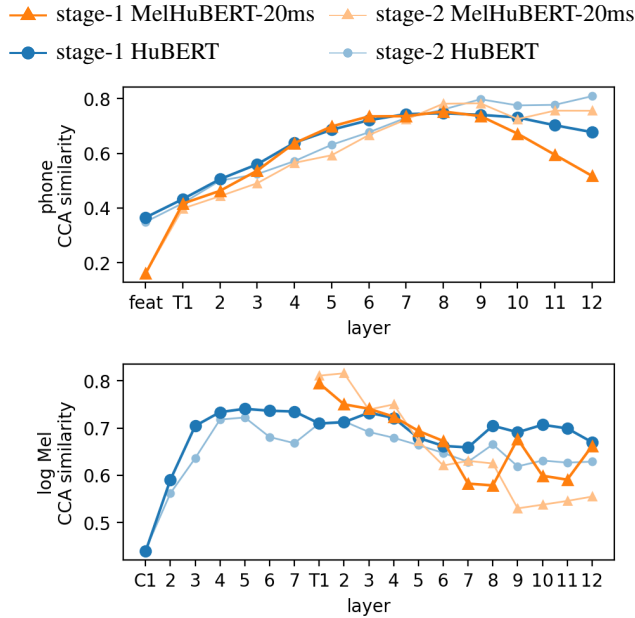
\begin{figure}
\centering
\definecolor{HuBERT-stg1}{HTML}{1F77B4}
\definecolor{HuBERT-stg2}{HTML}{8FBBD9}
\definecolor{MelHuBERT-stg1}{HTML}{FF7F0E}
\definecolor{MelHuBERT-stg2}{HTML}{FFBF86}
\begin{tikzpicture}
  \draw[MelHuBERT-stg1, thick] (0, 0) edge node[regular polygon, regular polygon sides=3, fill=MelHuBERT-stg1, inner sep=1.25pt] {} (0.5, 0);
  \node[right, font=\small] at (0.5, 0) {stage-1 MelHuBERT-20ms};
  \draw[MelHuBERT-stg2] (4.25, 0) edge node[regular polygon, regular polygon sides=3, fill=MelHuBERT-stg2, inner sep=1pt] {} (4.75, 0);
  \node[right, font=\small] at (4.75, 0) {stage-2 MelHuBERT-20ms};
  \draw[HuBERT-stg1, thick] (0, -0.5) edge node[circle, fill=HuBERT-stg1, inner sep=1.75pt] {} (0.5, -0.5);
  \node[right, font=\small] at (0.5, -0.5) {stage-1 HuBERT};
  \draw[HuBERT-stg2] (4.25, -0.5) edge node[circle, fill=HuBERT-stg2, inner sep=1.25pt] {} (4.75, -0.5);
  \node[right, font=\small] at (4.75, -0.5) {stage-2 HuBERT};
\end{tikzpicture}
\\
\includegraphics[width=8cm]{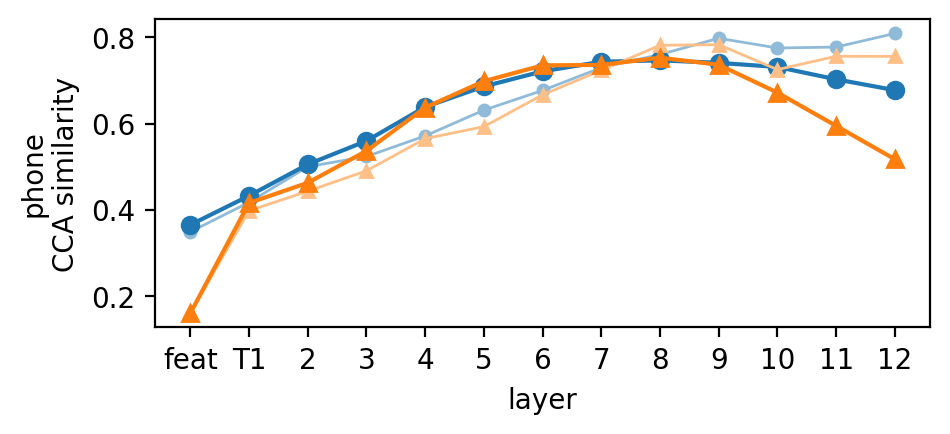} \\
\includegraphics[width=8cm]{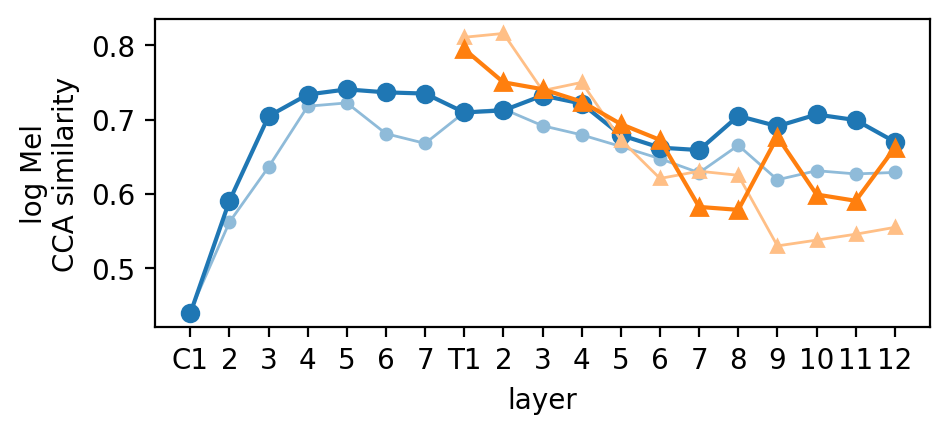}
\caption{\emph{Top:} CCA similarity between different layers and phones. \emph{Bottom:} CCA similarity between different layers and log Mel spectrograms. \texttt{C1} indicates the first convolution layer, \texttt{T1} indicates the first Transformer layer, and \texttt{feat} is the input to \texttt{T1}. The models are pre-trained on LibriSpeech 360-hour subset. \label{fig:cca_layerwise}}
\end{figure}

The similarity to Mel spectrograms is shown in the bottom plot of Figure \ref{fig:cca_layerwise}.
Similar to the findings in \cite{pasad2023comparative}, convolution layers in HuBERT have high similarity to Mel spectrograms.
The similarity stays about the same throughout the Transformer
layers in HuBERT, but in MelHuBERT, the similarity decreases until layer eleventh and increases again in the twelfth layer.
This is again consistent with the autocoding behavior of pre-training.

Compared to stage 1, similarity to Mel spectrograms is lower in stage 2.
The decrease in similarity is more significant for MelHuBERT, especially in higher layers.
We again attribute this to the targets (for stage-2 pre-training) being more phonetically prominent.

\subsection{Strengths of HuBERT's convolution layers}

Based on the layer-wise analysis, convolution layers
have a high similarity to Mel spectrograms.
Since convolution layers are trained, we further study whether
there is information more accessible for convolution layers
than for Mel spectrograms.
We probe phonetic information with ASR on both Mel spectrograms and the convolution layers following SUPERB (with a two-layer LSTM and CTC \cite{graves2006connectionist}).
In addition, we probe the fundamental frequency ($F_0$), following SUPERB-prosody \cite{lin2023utility}. 
The reference pitch is computed with pYAAPT and we train linear regression models to predict
the log of $F_0$.

The results of probing are shown in Table \ref{tbl:cmp_logmel_hubertconv}.
Mel spectrograms perform slightly better than convolution layers on ASR,
while convolution layers are better at extracting $F_0$.
Interestingly, despite HuBERT using the sixth layer as targets in stage-2 pre-training,
the ability to track $F_0$ improves over stage 1.
Technically, it is possible to count the harmonics in Mel spectrograms
to determine $F_0$, but $F_0$ is less \emph{linearly} accessible due to the nonlinear
Mel scale.

\begin{table}
  \caption{A comparison of Mel spectrograms and HuBERT's convolution layers on ASR and $F_0$ tracking.
    \texttt{C7} denotes the seventh (and last) convolution layers in HuBERT.
    \label{tbl:cmp_logmel_hubertconv}}
  \centering
  \vspace{1em}
  \begin{tabular}{ccc}
  \toprule
                \multirow{2}{*}{Feature} & ASR & $\log F_0$ \\
                      & (WER) & (MSE)\\
  \midrule
  Mel spectrograms    & \textbf{23.18}  & 0.089 \\
  stage-1 HuBERT-C7   & 24.97 & 0.024 \\
  stage-2 HuBERT-C7   & 24.02 & \textbf{0.021} \\
  \bottomrule
  \end{tabular}
  \vspace{-1em}
\end{table}

\subsection{Strengths of using Mel spectrograms}

\begin{table}
  \caption{A comparison of MelHuBERT and HuBERT pre-trained on the 100-hour subset of LibriSpeech
  in terms of PR, SID, and ASR. 
  \label{tbl:100-hour}}
  \vspace{1em}
  \centering
  \begin{tabular}{ccccc}
  \toprule
  \multirow{2}{*}{Model} & PR & SID & ASR \\
  & (PER) & (ER) & (WER) \\
  \midrule  
  MelHuBERT-20ms & \textbf{19.2} & \textbf{38.9} & \textbf{17.4} \\
  HuBERT  & 32.0 & 43.7 & 19.4 \\
  \bottomrule
  \end{tabular}
\end{table}

Since MelHuBERT does not require to learn a feature extractor from wave samples,
we suspect MelHuBERT would have a stronger edge over HuBERT
when there is not that much data for pre-training.
To study this, we pre-train HuBERT and MelHuBERT on the 100-hour subset of LibriSpeech (about a third of the pre-training data for the experiments before).
The results are shown in Table \ref{tbl:100-hour}.
Indeed, MelHuBERT tops HuBERT in all PR, SID and ASR.
We confirm that MelHuBERT has an advantage over HuBERT in low-resource settings,
in terms both data and compute.

\subsection{The pre-training speed-up of MelHuBERT}

\begin{figure}
    \centering
    \includegraphics[width=8cm]{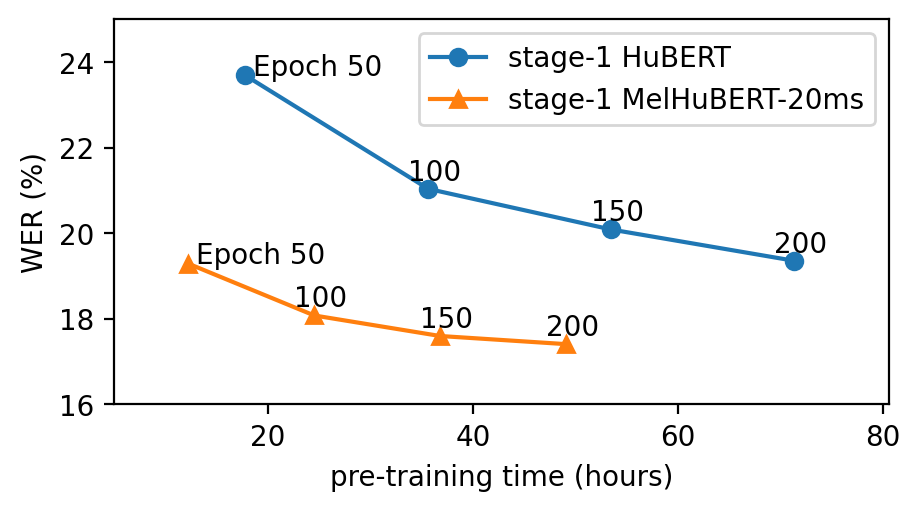}
    \vspace{-1.4em}
    \caption{Pre-raining time required to epoch 50, 100, 150, and 200
      on the 100-hour subset of LibriSpeech,
      and the respective downstream ASR performance.
      \label{fig:wer_time}}
    \vspace{-0.8em}
\end{figure}

To show the speed-up of MelHuBERT against HuBERT,
we measure the training time to epoch 50, 100, 150, and 200
when pre-training on the 100-hour subset of LibriSpeech and the respective downstream ASR performance. 
The experiment is conducted on a single NVIDIA RTX 3090 without mixed precision training.
The results are shown in Figure \ref{fig:wer_time}.
MelHuBERT-20ms takes about 14.7 minutes per epoch, while HuBERT takes about 21.4 minutes.
The result shows that our training recipe saves 31.2\% of the pre-training time.
MelHuBERT is not only faster to train but also enjoys a lower WER.

\section{Related work}

MPC proposed by \cite{jiang2019improving} is the first to learn speech representations with 12-layer Transformers and masked prediction on Mel spectrograms.
Subsequent variants of MPC include DeCoAR 2.0 \cite{ling2020decoar}, Mockingjay \cite{liu2020mockingjay}, and Tera \cite{liu2021tera}.
However, MelHuBERT outperforms all of them on the SUPERB benchmark, despite only pre-trained
on a third of LibriSpeech.\footnote{The downstream performance of DeCoAR 2.0, Mockingjay, and Tera can be found at https://superbbenchmark.org/leaderboard.}

BEST-RQ \cite{chiu2022self} also learns speech representations on Mel spectrograms,
but it makes other changes to HuBERT, trading the initial $k$-means clustering with clever initialization.
BEST-RQ is only evaluated on ASR with fine-tuning, and does not explore stage-2 pre-training. 

Our motivation is similar to \cite{william2023ruducing}, with the goal of reducing the barrier of self-supervised learning under limited computational resources. However, they focus on reproducing HuBERT without altering the model architecture and training objective, while we study the impact of model architecture and several key components in pre-training. 
MelHuBERT not only runs faster but also requires less memory consumption, amenable to training on a single consumer-grade GPU (RTX 3090), while they use 8 high-end GPUs (A100).

Reducing computational cost by replacing convolution layers in wav2vec 2.0 is explored in \cite{parcollet2023efficiency}.
Similar to BEST-RQ, they only evaluate their models on ASR with fine-tuning,
while we include an analyses on the similarities and differences of convolution layers
and Mel spectrograms.
Their models are based on wav2vec 2.0 and do not include stage-2 pre-training,
while we study stage-2 pre-training with layer-wise analysis and downstream tasks.

The experiment of replacing the convolution layers of pretrained HuBERT with Mel spectrograms by a front-end adapter has been explored in \cite{chen2023front}. Their model yields a comparable WER when compared to the original HuBERT. Again, they only evaluate their model on ASR and do not train their model from scratch. Additionally, we have a more in-depth analysis between these two kinds of features.

\section{Conclusion}

In this paper, we propose MelHuBERT, a simplified version of HuBERT 
on both model architecture and training recipe.
MelHuBERT achieves favorable performance against HuBERT and is more efficient on both pre-training and inference, with 31.2\% reduction on pre-training time and 33.5\% reduction on MACs per one second speech.
A comprehensive analysis is conducted betweeen HuBERT and MelHuBERT, 
including the differences of the learned representations, 
the strengths of each model, 
and the pre-training efficiency.
There are still many open questions in self-supervised learning,
and the simplified training recipe could facilitate
studies on the interaction between self-supervised learning and speech representations,
beyond performance on downstream tasks.

\bibliographystyle{IEEEbib}
\bibliography{compression}

\begin{thebibliography}{10}

\bibitem{chung2019unsupervised}
Yu-An Chung, Wei-Ning Hsu, Hao Tang, and James Glass,
\newblock ``An unsupervised autoregressive model for speech representation
  learning,''
\newblock in {\em Interspeech}, 2019.

\bibitem{schneider2019wav2vec}
Steffen Schneider, Alexei Baevski, Ronan Collobert, and Michael Auli,
\newblock ``{wav2vec}: Unsupervised pre-training for speech recognition,''
\newblock {\em arXiv preprint arXiv:1904.05862}, 2019.

\bibitem{baevski2020wav2vec}
Alexei Baevski, Yuhao Zhou, Abdelrahman Mohamed, and Michael Auli,
\newblock ``wav2vec 2.0: A framework for self-supervised learning of speech
  representations,''
\newblock {\em NeurIPS}, 2020.

\bibitem{chung2020vector}
Yu-An Chung, Hao Tang, and James Glass,
\newblock ``Vector-quantized autoregressive predictive coding,''
\newblock in {\em Interspeech}, 2020.

\bibitem{hsu2021hubert}
Wei-Ning Hsu, Benjamin Bolte, Yao-Hung~Hubert Tsai, Kushal Lakhotia, Ruslan
  Salakhutdinov, and Abdelrahman Mohamed,
\newblock ``{HuBERT}: Self-supervised speech representation learning by masked
  prediction of hidden units,''
\newblock {\em IEEE/ACM Transctions on Audio, Speech and Language Processing},
  2021.

\bibitem{chen2022wavlm}
Sanyuan Chen, Chengyi Wang, Zhengyang Chen, Yu~Wu, Shujie Liu, Zhuo Chen, Jinyu
  Li, Naoyuki Kanda, Takuya Yoshioka, Xiong Xiao, et~al.,
\newblock ``{WavLM}: Large-scale self-supervised pre-training for full stack
  speech processing,''
\newblock {\em IEEE Journal of Selected Topics in Signal Processing}, 2022.

\bibitem{huang2022masked}
Po-Yao Huang, Hu~Xu, Juncheng Li, Alexei Baevski, Michael Auli, Wojciech
  Galuba, Florian Metze, and Christoph Feichtenhofer,
\newblock ``Masked autoencoders that listen,''
\newblock {\em NeurIPS}, 2022.

\bibitem{zhang2022bigssl}
Yu~Zhang, Daniel~S Park, Wei Han, James Qin, Anmol Gulati, Joel Shor, Aren
  Jansen, Yuanzhong Xu, Yanping Huang, Shibo Wang, et~al.,
\newblock ``{BigSSL}: Exploring the frontier of large-scale semi-supervised
  learning for automatic speech recognition,''
\newblock {\em IEEE Journal of Selected Topics in Signal Processing}, 2022.

\bibitem{jiang2019improving}
Dongwei Jiang, Xiaoning Lei, Wubo Li, Ne~Luo, Yuxuan Hu, Wei Zou, and Xiangang
  Li,
\newblock ``Improving transformer-based speech recognition using unsupervised
  pre-training,''
\newblock {\em arXiv preprint arXiv:1910.09932}, 2019.

\bibitem{ling2020decoar}
Shaoshi Ling and Yuzong Liu,
\newblock ``{DeCoAR 2.0}: Deep contextualized acoustic representations with
  vector quantization,''
\newblock {\em arXiv preprint arXiv:2012.06659}, 2020.

\bibitem{baade2022mae}
Alan Baade, Puyuan Peng, and David Harwath,
\newblock ``{MAE-AST}: Masked autoencoding audio spectrogram transformer,''
\newblock {\em arXiv preprint arXiv:2203.16691}, 2022.

\bibitem{chong2023masked}
Dading Chong, Helin Wang, Peilin Zhou, and Qingcheng Zeng,
\newblock ``Masked spectrogram prediction for self-supervised audio
  pre-training,''
\newblock in {\em ICASSP}, 2023.

\bibitem{sainath2015learning}
Tara Sainath, Ron~J. Weiss, Kevin Wilson, Andrew~W. Senior, and Oriol Vinyals,
\newblock ``Learning the speech front-end with raw waveform {CLDNNs},''
\newblock in {\em Interspeech}, 2015.

\bibitem{ravanelli2018speaker}
Mirco Ravanelli and Yoshua Bengio,
\newblock ``Speaker recognition from raw waveform with {SincNet},''
\newblock in {\em SLT}, 2018.

\bibitem{oord2018representation}
Aaron van~den Oord, Yazhe Li, and Oriol Vinyals,
\newblock ``Representation learning with contrastive predictive coding,''
\newblock {\em arXiv preprint arXiv:1807.03748}, 2018.

\bibitem{gutmann2010noise}
Michael Gutmann and Aapo Hyv{\"a}rinen,
\newblock ``Noise-contrastive estimation: A new estimation principle for
  unnormalized statistical models,''
\newblock in {\em AISTAT}, 2010.

\bibitem{yang2022autoregressive}
Gene-Ping Yang, Sung-Lin Yeh, Yu-An Chung, James Glass, and Hao Tang,
\newblock ``Autoregressive predictive coding: A comprehensive study,''
\newblock {\em IEEE Journal of Selected Topics in Signal Processing}, 2022.

\bibitem{yeh2022autoregressive}
Sung-Lin Yeh and Hao Tang,
\newblock ``{Autoregressive Co-Training for Learning Discrete Speech
  Representations},''
\newblock in {\em Interspeech}, 2022.

\bibitem{baevski2019vq}
Alexei Baevski, Steffen Schneider, and Michael Auli,
\newblock ``{vq-wav2vec}: Self-supervised learning of discrete speech
  representations,''
\newblock {\em arXiv preprint arXiv:1910.05453}, 2019.

\bibitem{fan2019reducing}
Angela Fan, Edouard Grave, and Armand Joulin,
\newblock ``Reducing transformer depth on demand with structured dropout,''
\newblock {\em arXiv preprint arXiv:1909.11556}, 2019.

\bibitem{micikevicius2017mixed}
Paulius Micikevicius, Sharan Narang, Jonah Alben, Gregory Diamos, Erich Elsen,
  David Garcia, Boris Ginsburg, Michael Houston, Oleksii Kuchaiev, Ganesh
  Venkatesh, et~al.,
\newblock ``Mixed precision training,''
\newblock {\em ICLR}, 2018.

\bibitem{ott2019fairseq}
Myle Ott, Sergey Edunov, Alexei Baevski, Angela Fan, Sam Gross, Nathan Ng,
  David Grangier, and Michael Auli,
\newblock ``{Fairseq}: A fast, extensible toolkit for sequence modeling,''
\newblock in {\em Proceedings of NAACL-HLT 2019: Demonstrations}, 2019.

\bibitem{shu2021superb}
Shu-wen Yang, Po-Han Chi, Yung-Sung Chuang, Cheng-I~Jeff Lai, Kushal Lakhotia,
  Yist~Y. Lin, Andy~T. Liu, Jiatong Shi, Xuankai Chang, Guan-Ting Lin, et~al.,
\newblock ``{SUPERB: Speech Processing Universal PERformance Benchmark},''
\newblock in {\em Interspeech}, 2021.

\bibitem{meng2022compressing}
Yen Meng, Hsuan-Jui Chen, Jiatong Shi, Shinji Watanabe, Paola Garcia, Hung-yi
  Lee, and Hao Tang,
\newblock ``On compressing sequences for self-supervised speech models,''
\newblock in {\em SLT}, 2022.

\bibitem{wu2022performance}
Felix Wu, Kwangyoun Kim, Jing Pan, Kyu~J Han, Kilian~Q. Weinberger, and Yoav
  Artzi,
\newblock ``Performance-efficiency trade-offs in unsupervised pre-training for
  speech recognition,''
\newblock in {\em ICASSP}, 2022.

\bibitem{peng2023structured}
Yifan Peng, Kwangyoun Kim, Felix Wu, Prashant Sridhar, and Shinji Watanabe,
\newblock ``Structured pruning of self-supervised pre-trained models for speech
  recognition and understanding,''
\newblock in {\em ICASSP}, 2023.

\bibitem{peng2023dphubert}
Yifan Peng, Yui Sudo, Shakeel Muhammad, and Shinji Watanabe,
\newblock ``{DPHuBERT}: Joint distillation and pruning of self-supervised
  speech models,''
\newblock in {\em Interspeech}, 2023.

\bibitem{parcollet2023efficiency}
Titouan Parcollet, Shucong Zhang, Rogier van Dalen, Alberto Gil C.~P. Ramos,
  and Sourav Bhattacharya,
\newblock ``On the (in)efficiency of acoustic feature extractors for
  self-supervised speech representation learning,''
\newblock in {\em Interspeech}, 2023.

\bibitem{chen2023front}
Xie Chen, Ziyang Ma, Changli Tang, Yujin Wang, and Zhisheng Zheng,
\newblock ``Front-end adapter: Adapting front-end input of speech based
  self-supervised learning for speech recognition,''
\newblock in {\em ICASSP}, 2023.

\bibitem{liu2020mockingjay}
Andy~T. Liu, Shu-wen Yang, Po-Han Chi, Po-chun Hsu, and Hung-yi Lee,
\newblock ``Mockingjay: Unsupervised speech representation learning with deep
  bidirectional transformer encoders,''
\newblock in {\em ICASSP}, 2020.

\bibitem{liu2021tera}
Andy~T. Liu, Shang-Wen Li, and Hung-yi Lee,
\newblock ``{TERA}: Self-supervised learning of transformer encoder
  representation for speech,''
\newblock {\em IEEE/ACM Transctions on Audio, Speech and Language Processing},
  2021.

\bibitem{gong2022ssast}
Yuan Gong, Cheng-I Lai, Yu-An Chung, and James Glass,
\newblock ``{SSAST}: Self-supervised audio spectrogram transformer,''
\newblock {\em AAAI}, 2022.

\bibitem{mcauliffe2017montreal}
Michael McAuliffe, Michaela Socolof, Sarah Mihuc, Michael Wagner, and Morgan
  Sonderegger,
\newblock ``{Montreal Forced Aligner: Trainable Text-Speech Alignment Using
  Kaldi.},''
\newblock in {\em Interspeech}, 2017.

\bibitem{pasad2021layer}
Ankita Pasad, Ju-Chieh Chou, and Karen Livescu,
\newblock ``Layer-wise analysis of a self-supervised speech representation
  model,''
\newblock in {\em ASRU}, 2021.

\bibitem{pasad2023comparative}
Ankita Pasad, Bowen Shi, and Karen Livescu,
\newblock ``Comparative layer-wise analysis of self-supervised speech models,''
\newblock in {\em ICASSP}, 2023.

\bibitem{graves2006connectionist}
Alex Graves, Santiago Fern{\'a}ndez, Faustino Gomez, and J{\"u}rgen
  Schmidhuber,
\newblock ``Connectionist temporal classification: labelling unsegmented
  sequence data with recurrent neural networks,''
\newblock {\em ICML}, 2006.

\bibitem{lin2023utility}
Guan-Ting Lin, Chi-Luen Feng, Wei-Ping Huang, Yuan Tseng, Tzu-Han Lin, Chen-An
  Li, Hung-yi Lee, and Nigel~G Ward,
\newblock ``On the utility of self-supervised models for prosody-related
  tasks,''
\newblock in {\em SLT}, 2023.

\bibitem{chiu2022self}
Chung-Cheng Chiu, James Qin, Yu~Zhang, Jiahui Yu, and Yonghui Wu,
\newblock ``Self-supervised learning with random-projection quantizer for
  speech recognition,''
\newblock {\em ICML}, 2022.

\bibitem{william2023ruducing}
William Chen, Xuankai Chang, Yifan Peng, Zhaoheng Ni, Soumi Maiti, and Shinji
  Watanabe,
\newblock ``{Reducing Barriers to Self-Supervised Learning: HuBERT Pre-training
  with Academic Compute},''
\newblock in {\em Interspeech}, 2023.

\end{thebibliography}

\end{document}